\documentclass{article}
\usepackage{spconf,amsmath,graphicx,hyperref}
\usepackage{graphics}
\usepackage{epsfig}
\usepackage{mathptmx}
\usepackage{times}
\usepackage{amsmath}
\usepackage{amssymb}
\usepackage{booktabs}
\usepackage{multirow}
\usepackage{url}
\usepackage{graphicx}
\usepackage{subcaption}
\usepackage{amsmath,amssymb,amsfonts}
\usepackage{float}
\usepackage{tikz}
\usepackage{pgfplots}
\usepackage{algorithm}
\usepackage{algorithmic}

\usepackage{setspace}

\usetikzlibrary{shapes,arrows,positioning,fit,backgrounds,decorations.pathreplacing,decorations.pathmorphing,calc,intersections,patterns,shadows}
\usepgfplotslibrary{fillbetween}
\pgfplotsset{compat=1.17}

\usepackage{graphics} 
\usepackage{epsfig} 
\usepackage{mathptmx} 
\usepackage{times} 
\usepackage{amsmath} 
\usepackage{amssymb}  
\usepackage{booktabs}
\usepackage{multirow}
\usepackage{url}
\usepackage{graphicx}
\usepackage{subcaption}
\usepackage{amsmath,amssymb,amsfonts}
\usepackage{float} 
\usepackage{tikz}
\usepackage{pgfplots}
\usetikzlibrary{arrows.meta, shapes,arrows,positioning,fit,backgrounds,decorations.pathreplacing,calc,patterns,shadows}

\newtheorem{theorem}{Theorem} 
\newtheorem{lemma}{Lemma}

\usepgfplotslibrary{groupplots}
\definecolor{NavyBlue}{RGB}{0,64,128}
\definecolor{Maroon}{RGB}{128,0,0}
\usepackage{xcolor}
\definecolor{OliveGreen}{RGB}{85,107,47}
\definecolor{Purple}{RGB}{128,0,128}
\definecolor{Gray}{RGB}{128,128,128}

\definecolor{primaryblue}{RGB}{41,74,112}
\definecolor{secondaryred}{RGB}{167,59,56}
\definecolor{tertiarygreen}{RGB}{58,101,77}
\definecolor{accentpurple}{RGB}{120,81,122}
\definecolor{accentorange}{RGB}{210,105,30}
\definecolor{lightgray}{RGB}{240,240,240}
\definecolor{darkgray}{RGB}{64,64,64}

\definecolor{techblue}{RGB}{52, 116, 186}   
\definecolor{techgreen}{RGB}{85, 168, 104}  
\definecolor{techred}{RGB}{196, 78, 82}     
\definecolor{techpurple}{RGB}{129, 114, 179}
\definecolor{techgray}{RGB}{77, 77, 77}     
\definecolor{bggray}{RGB}{242, 242, 242}    

\definecolor{myblue}{RGB}{235, 245, 251}   
\definecolor{mybluedark}{RGB}{40, 116, 166} 
\definecolor{mygreen}{RGB}{233, 247, 239}  
\definecolor{mygreendark}{RGB}{35, 155, 86} 
\definecolor{mypurple}{RGB}{244, 236, 247} 
\definecolor{mypurpledark}{RGB}{125, 60, 152} 
\definecolor{myorange}{RGB}{254, 245, 231} 
\definecolor{myorangedark}{RGB}{211, 84, 0} 
\definecolor{mygray}{RGB}{80, 80, 80}      

\def\x{{\mathbf x}}

\title{AWGformer: Adaptive Wavelet-Guided Transformer for Multi-Resolution Time Series Forecasting}
\name{Wei Li\thanks{ORCID: \href{https://orcid.org/0009-0008-8108-4854}{0009-0008-8108-4854}, E-mail: \href{mailto:liwei008009@163.com}{liwei008009@163.com}.}}
\address{School of Computer Engineering and Science, Shanghai University, Shanghai, China}
\begin{document}
%
\maketitle
\begin{abstract}
Time series forecasting requires capturing patterns across multiple temporal scales while maintaining computational efficiency. This paper introduces AWGformer, a novel architecture that integrates adaptive wavelet decomposition with cross-scale attention mechanisms for enhanced multi-variate time series prediction. Our approach comprises: (1) an Adaptive Wavelet Decomposition Module (AWDM) that dynamically selects optimal wavelet bases and decomposition levels based on signal characteristics; (2) a Cross-Scale Feature Fusion (CSFF) mechanism that captures interactions between different frequency bands through learnable coupling matrices; (3) a Frequency-Aware Multi-Head Attention (FAMA) module that weights attention heads according to their frequency selectivity; (4) a Hierarchical Prediction Network (HPN) that generates forecasts at multiple resolutions before reconstruction. Extensive experiments on benchmark datasets demonstrate that AWGformer achieves significant average improvements over state-of-the-art methods, with particular effectiveness on multi-scale and non-stationary time series. Theoretical analysis provides convergence guarantees and establishes the connection between our wavelet-guided attention and classical signal processing principles.
\end{abstract}
\vspace{-0.2cm}
	\section{Introduction}
\vspace{-0.2cm}
Time series forecasting is fundamental to numerous applications including climate modeling \cite{reichstein2019deep}, labor markets \cite{liu2026health}, and driving and traffic management \cite{zeng2025FSDrive}. The inherent multi-scale nature of real-world time series—from high-frequency noise to long-term trends—poses significant challenges for accurate prediction. Traditional approaches like ARIMA \cite{box2015time} and exponential smoothing struggle with complex multi-scale patterns. Deep learning methods, including temporal convolutional networks \cite{bai2018empirical}, and recent transformer variants \cite{zhou2021informer,wu2021autoformer,nie2023time}, have shown promise but often treat all temporal scales uniformly, missing opportunities for scale-specific modeling. Wavelet analysis provides a natural framework for multi-resolution signal decomposition \cite{mallat1999wavelet}, successfully applied in signal denoising \cite{donoho1994ideal} and compression \cite{antonini1992image}. Despite these advances, a unified framework that end-to-end learns the wavelet decomposition within a Transformer architecture remains an open challenge.

We propose AWGformer, a novel architecture bridging classical wavelet theory and transformer-based forecasting. Our main contributions are summarized as follows:
\begin{itemize}\setlength{\itemsep}{0pt}\setlength{\parsep}{0pt}\setlength{\parskip}{0pt}
	\item \textbf{Adaptive Wavelet Decomposition (AWDM):} A learnable module that dynamically adjusts wavelet bases and decomposition levels to match signal characteristics.
	\item \textbf{Cross-Scale Feature Fusion (CSFF):} A mechanism utilizing learnable coupling matrices to capture non-linear interactions between different frequency bands.
	\item \textbf{Frequency-Aware Attention (FAMA):} A specialized attention design where heads are constrained by frequency selectivity.
	\item \textbf{Comprehensive Evaluation:} Theoretical guarantees on convergence and extensive experiments demonstrating state-of-the-art performance on benchmarks.
\end{itemize}

\vspace{-0.2cm}

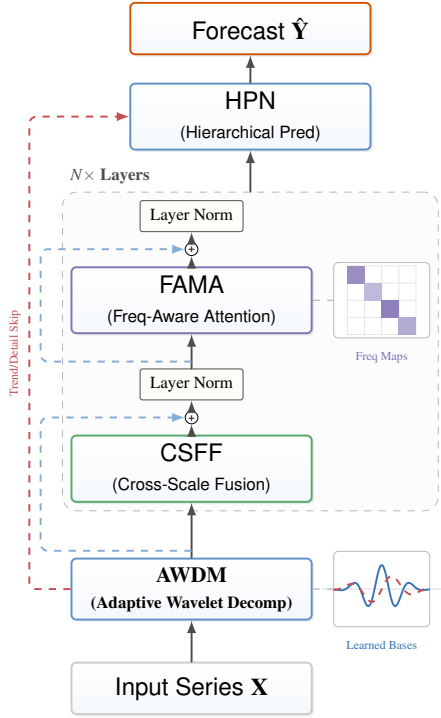
\begin{figure}[t]
	\centering
	\resizebox{0.7\columnwidth}{!}{
		\begin{tikzpicture}[
			node distance=0.4cm and 0.4cm, 
			every node/.style={font=\sffamily},
			main_block/.style={
				rectangle, 
				rounded corners=2pt, 
				draw=techblue!80, 
				thick,
				top color=white,
				bottom color=techblue!5,
				minimum width=3.5cm, 
				minimum height=0.75cm,
				align=center,
				drop shadow={opacity=0.15, shadow xshift=1pt, shadow yshift=-1pt}
			},
			viz_box/.style={
				rectangle,
				rounded corners=2pt,
				draw=techgray!40,
				fill=white,
				minimum width=1.4cm, 
				minimum height=1.1cm,
				inner sep=0pt
			},
			norm_block/.style={
				rectangle,
				draw=techgray,
				fill=myorange!20,
				rounded corners=1pt,
				font=\scriptsize,
				minimum width=1.5cm,
				minimum height=0.3cm
			},
			flow/.style={-latex, thick, draw=techgray, line cap=round},
			skip_line/.style={-latex, thick, dashed, draw=techblue!60, rounded corners=3pt}
			]
			
			\node (input) [main_block, fill=gray!10, draw=gray!40] {Input Series $\mathbf{X}$};
			
			\node (awdm) [main_block, above=0.6cm of input, font=\bfseries\small] {AWDM\\ \scriptsize (Adaptive Wavelet Decomp)};
			\draw[flow] (input) -- (awdm);
			
			\node (awdm_viz) [viz_box, right=0.3cm of awdm, label={[font=\tiny, text=techblue, yshift=-2pt]below:Learned Bases}] {};
			\begin{scope}[shift={(awdm_viz.center)}, scale=0.35]
				\draw[techgray!20] (-2.5,0) -- (2.5,0);
				\draw[techblue, thick, smooth] plot[domain=-2:2, samples=40] (\x, {cos(300*\x)*exp(-\x*\x)});
				\draw[techred, thick, dashed, smooth] plot[domain=-2:2, samples=40] (\x, {0.6*sin(200*\x)*exp(-0.6*\x*\x)});
			\end{scope}
			\draw[dashed, techgray!50] (awdm.east) -- (awdm_viz.west);
			
			\coordinate (enc_start) at ($(awdm.north)+(0,0.5)$);
			
			\node (csff) [main_block, above=0.8cm of awdm, fill=techgreen!5, draw=techgreen] {CSFF\\ \scriptsize (Cross-Scale Fusion)};
			
			\node (add1) [circle, draw, inner sep=0.5pt, above=0.15cm of csff] {\tiny +};
			\node (norm1) [norm_block, above=0.15cm of add1] {Layer Norm};
			
			\draw[flow] (awdm) -- (csff);
			\draw[flow] (csff) -- (add1);
			\draw[flow] (add1) -- (norm1);
			\draw[skip_line] ($(awdm.north)+(0,0.1)$) -- ++(-2.2,0) |- (add1.west);
			
			\node (fama) [main_block, above=0.5cm of norm1, fill=techpurple!5, draw=techpurple] {FAMA\\ \scriptsize (Freq-Aware Attention)};
			
			\node (fama_viz) [viz_box, right=0.3cm of fama, label={[font=\tiny, text=techpurple, yshift=-2pt]below:Freq Maps}] {};
			\begin{scope}[shift={($(fama_viz.center)+(-0.5,-0.5)$)}, scale=0.25]
				\draw[step=1cm, gray!20, very thin] (0,0) grid (4,4);
				\fill[techpurple!80] (0,3) rectangle (1,4);
				\fill[techpurple!50] (1,2) rectangle (2,3);
				\fill[techpurple!90] (2,1) rectangle (3,2);
				\fill[techpurple!60] (3,0) rectangle (4,1);
			\end{scope}
			\draw[dashed, techgray!50] (fama.east) -- (fama_viz.west);
			
			\node (add2) [circle, draw, inner sep=0.5pt, above=0.15cm of fama] {\tiny +};
			\node (norm2) [norm_block, above=0.15cm of add2] {Layer Norm};
			
			\draw[flow] (norm1) -- (fama);
			\draw[flow] (fama) -- (add2);
			\draw[flow] (add2) -- (norm2);
			\draw[skip_line] ($(norm1.north)+(0,0.1)$) -- ++(-2.2,0) |- (add2.west);
			
			\begin{scope}[on background layer]
				\node [fit=(csff)(norm2)(fama_viz)(add1), draw=techgray!40, dashed, rounded corners=6pt, fill=gray!2] (enc_box) {};
				\node [anchor=south west, font=\bfseries\scriptsize, text=techgray] at (enc_box.north west) {$N \times$ Layers};
			\end{scope}
			
			\node (hpn) [main_block, above=0.6cm of enc_box, fill=techgray!10] {HPN\\ \scriptsize (Hierarchical Pred)};
			\node (output) [main_block, above=0.4cm of hpn, draw=myorangedark, fill=myorangedark!5] {Forecast $\hat{\mathbf{Y}}$};
			
			\draw[flow] (enc_box.north -| hpn.south) -- (hpn.south);
			\draw[flow] (hpn) -- (output);
			
			\draw[skip_line, color=techred] (awdm.west) -- ++(-0.6,0) |- node[pos=0.25, rotate=90, above, font=\tiny, text=techred]{Trend/Detail Skip} (hpn.west);
			
		\end{tikzpicture}
	}
	\caption{The AWGformer architecture. The model integrates (1) \textbf{AWDM} for adaptive decomposition, (2) Stacked encoder layers with \textbf{CSFF} and \textbf{FAMA} for cross-scale modeling, and (3) \textbf{HPN} for hierarchical forecasting. Visualizations show learned wavelets and frequency-aware attention maps.}
	\label{fig:architecture}
\end{figure}

\vspace{-0.2cm}
\section{Related Work}
\vspace{-0.2cm}
Deep learning has revolutionized time series forecasting. Transformer-based models have shown particular promise: Informer \cite{zhou2021informer} uses ProbSparse attention for efficiency, Autoformer \cite{wu2021autoformer} incorporates decomposition, and PatchTST \cite{nie2023time} applies patching strategies. Attention mechanisms have become a dominant paradigm in sequence modeling, demonstrating success in diverse domains including traffic prediction~\cite{song2025smartcity}. Recent work like TimesNet \cite{wu2023timesnet} explores 2D variations for periodicity modeling. However, these methods typically operate in a single domain without explicit multi-resolution modeling.
Wavelets provide natural multi-resolution representations \cite{mallat1999wavelet}. Applications include WaveNet \cite{oord2016wavenet} for audio generation and wavelet pooling \cite{williams2018wavelet} for CNNs. Recent work combines wavelets with neural networks for denoising \cite{liu2020waveletnet}. However, most approaches use fixed wavelet bases rather than learning adaptive decompositions.
Multi-scale approaches include hierarchical models \cite{rangapuram2018deep} and frequency-domain methods \cite{zhou2022fedformer}. iTransformer \cite{liu2024itransformer} is another novel method. The importance of structural decomposition has been evidenced in recent works like JanusVLN~\cite{zeng2025janusvln}, which validates the benefits of decoupling semantics from spatiality. Similarly, diffusion-based architectures~\cite{li2025timeflowdiffuser} have achieved remarkable performance by modeling temporal dynamics through generative processes. Distinct from these approaches, our work focuses on adaptively learning both wavelet decomposition and cross-scale interactions. This builds upon our preliminary exploration of wavelet-based time series modeling~\cite{li2025swift}, advancing it into a fully learnable framework. This philosophy aligns with broader trends in optimizing representations, as seen in recent advances in efficient coding~\cite{li2025preference} and medical imaging~\cite{li2025efficient}. Furthermore, while our model focuses on signal processing, the field is also evolving to tackle complex decision-making via Large Language Models~\cite{song2026decision} and to enhance trust through novel explainability frameworks~\cite{shen2025aienhanced}.
\vspace{-0.2cm}
\section{Methodology}
\vspace{-0.2cm}
\subsection{Problem Formulation}
Given a multivariate time series $\mathbf{X} = [\mathbf{x}_1, ..., \mathbf{x}_T] \in \mathbb{R}^{T \times D}$ with $T$ time steps and $D$ dimensions, we aim to predict future values $\mathbf{Y} = [\mathbf{x}_{T+1}, ..., \mathbf{x}_{T+H}] \in \mathbb{R}^{H \times D}$ for horizon $H$.

\subsection{Adaptive Wavelet Decomposition Module (AWDM)}

Traditional wavelet decomposition uses fixed basis functions. We propose learning adaptive wavelets tailored to the input signal characteristics.

\subsubsection{Learnable Wavelet Transform}
We parameterize the wavelet and scaling functions using neural networks:
$
	\psi_\theta(t) = \sigma(g_\theta(t)) \cdot \cos(\omega_\theta t + \phi_\theta)
$
where $g_\theta$ is a learnable envelope function, and $\omega_\theta, \phi_\theta$ are learnable frequency and phase parameters.

The multi-level decomposition is computed as:
$
	\mathbf{H}_j[n] = \sum_{k} \mathbf{x}[k] \psi_{j,n-k}
$
$
	\mathbf{L}_J[n] = \sum_{k} \mathbf{x}[k] \phi_{J,n-k}
$
where $\mathbf{H}_j$ are detail coefficients at level $j$, $\mathbf{L}_J$ is the approximation at the coarsest level $J$, and $\psi_{j,k} = 2^{-j/2}\psi_\theta(2^{-j}t - k)$.

\subsubsection{Adaptive Level Selection}
The optimal decomposition level is determined by:
$
	J^* = \arg\min_J \mathcal{L}_{\text{recon}}(J) + \lambda \mathcal{L}_{\text{sparse}}(J)
$
where $\mathcal{L}_{\text{recon}}$ measures reconstruction error and $\mathcal{L}_{\text{sparse}}$ encourages sparsity in the wavelet domain.

\subsection{Cross-Scale Feature Fusion (CSFF)}

Different frequency bands contain complementary information. We model their interactions through learnable coupling:

\begin{equation}
	\mathbf{F}_{\text{fused}} = \sum_{i,j} \mathbf{W}_{ij} \odot (\mathbf{H}_i \otimes \mathbf{H}_j)
\end{equation}

where $\mathbf{W}_{ij}$ are learnable coupling matrices and $\otimes$ denotes outer product.
To preserve scale-specific information, we apply residual connections:
$
	\mathbf{F}_j' = \mathbf{F}_j + \alpha_j \mathbf{F}_{\text{fused}}^{(j)}
$
where $\alpha_j$ are learnable gates controlling fusion strength.

To further stabilize training, we introduce a spectral dropout strategy: during each forward pass, we randomly zero out 20\% of the frequency channels in $\mathbf{F}_{\text{fused}}$.
This acts as a data-dependent regularizer, preventing the coupling matrices $\mathbf{W}_{ij}$ from overfitting to spurious cross-band correlations.
The dropout rate is annealed from 0.3 to 0.05 over the first 10\,k iterations.

\subsection{Frequency-Aware Multi-Head Attention (FAMA)}

Standard multi-head attention treats all frequencies equally. We propose frequency-aware attention where each head specializes in specific frequency bands.

\subsubsection{Frequency-Selective Heads}
Each attention head $h$ is associated with a frequency response function:

\begin{equation}
	H_h(\omega) = \exp\left(-\frac{(\omega - \omega_h)^2}{2\sigma_h^2}\right)
\end{equation}

where $\omega_h$ and $\sigma_h$ are learnable center frequency and bandwidth.
\vspace{-0.2cm}
\subsubsection{Weighted Attention}
\vspace{-0.2cm}
The attention computation is modified as:

\begin{equation}
	\text{Attention}_h(\mathbf{Q}, \mathbf{K}, \mathbf{V}) = \text{softmax}\left(\frac{\mathbf{Q}\mathbf{K}^T}{\sqrt{d_k}} \odot \mathbf{M}_h\right)\mathbf{V}
\end{equation}

where $\mathbf{M}_h$ is a frequency-dependent masking matrix derived from $H_h(\omega)$.
\vspace{-0.2cm}
\subsection{Hierarchical Prediction Network (HPN)}
\vspace{-0.2cm}
Instead of predicting directly in the time domain, we generate predictions at multiple resolutions:
$
	\hat{\mathbf{L}}_J = f_{\theta_L}(\mathbf{L}_J, \mathbf{F}_{\text{fused}})
$
$
	\hat{\mathbf{H}}_j = f_{\theta_j}(\mathbf{H}_j, \mathbf{F}_{\text{fused}})
$
The final prediction is obtained through inverse wavelet transform:
$
	\hat{\mathbf{Y}} = \mathcal{W}^{-1}(\hat{\mathbf{L}}_J, \{\hat{\mathbf{H}}_j\}_{j=1}^J)
$
\subsection{Training Objective}

The total loss combines prediction error, reconstruction quality, and regularization:
$
	\mathcal{L} = \mathcal{L}_{\text{pred}} + \lambda_1 \mathcal{L}_{\text{recon}} + \lambda_2 \mathcal{L}_{\text{ortho}} + \lambda_3 \mathcal{L}_{\text{smooth}}
$
where:
$\mathcal{L}_{\text{pred}} = \|\mathbf{Y} - \hat{\mathbf{Y}}\|_2^2$ is the forecasting error,
$\mathcal{L}_{\text{recon}} = \|\mathbf{X} - \mathcal{W}^{-1}(\mathcal{W}(\mathbf{X}))\|_2^2$ ensures perfect reconstruction,
$\mathcal{L}_{\text{ortho}} = \|\mathbf{\psi}^T\mathbf{\psi} - \mathbf{I}\|_F^2$ encourages orthogonal wavelets,
$\mathcal{L}_{\text{smooth}} = \sum_j \|\nabla^2 \psi_j\|_2^2$ promotes smooth basis functions.

\vspace{-0.2cm}
\subsection{Theoretical Properties}
\vspace{-0.2cm}
\begin{theorem}[Approximation Guarantee]
	For any $f \in L^2(\mathbb{R})$ with bounded variation, the adaptive wavelet decomposition with $J$ levels achieves:
	$
		\|f - f_J\|_2 \leq C \cdot 2^{-J\alpha} \|f\|_{BV}
	$
	where $\alpha$ depends on the smoothness of learned wavelets and $C$ is a constant.
\end{theorem}

\begin{lemma}[Frequency Localization]
	The frequency-aware attention heads achieve frequency selectivity with resolution:
	$
		\Delta f \cdot \Delta t \geq \frac{1}{4\pi}
	$
	satisfying the uncertainty principle.
\end{lemma}

\section{Experiments}

\subsection{Datasets and Setup}

We evaluate on benchmark datasets:
\textbf{ETT (4 subsets):} Electricity transformer temperature \cite{zhou2021informer};
\textbf{Traffic:} Road occupancy rates from California \cite{li2018diffusion};
\textbf{Electricity:} Hourly electricity consumption \cite{lai2018modeling};
We use standard train/validation/test splits and evaluate with MSE and MAE metrics across prediction horizons $\{96, 192, 336, 720\}$.
AWGformer uses 3 decomposition levels with learnable Daubechies-style wavelets.

\begin{table*}[t]
	\caption{Multivariate forecasting results. Best in \textbf{bold}, second best \underline{underlined}.}
	\label{tab:main_results}
	\centering
	\resizebox{0.7\textwidth}{!}{ 
		\begin{tabular}{l|cccc|cccc}
			\toprule
			\multirow{2}{*}{Method} & \multicolumn{4}{c|}{ETTh1 (MSE/MAE)} & \multicolumn{4}{c}{ETTh2 (MSE/MAE)} \\
			& 96 & 192 & 336 & 720 & 96 & 192 & 336 & 720 \\
			\midrule
			Autoformer 		& .449/.459 & .500/.482 & .521/.496 & .514/.512 & .346/.388 & .456/.452 & .482/.486 & .515/.511 \\
			FEDformer 		& \underline{.376}/.419 & \underline{.420}/.448 & \underline{.459}/.465 & .506/.507 & .358/.397 & .429/.439 & .496/.487 & .463/.474 \\
			TimesNet 		& .384/.402 & .436/\underline{.429} & .491/.469 & .521/.500 & .340/.374 & .402/.414 & .452/.452 & .462/.468 \\
			PatchTST 		& .414/.419 & .460/.445 & .501/.466 & \underline{.500}/\underline{.488} & .302/\underline{.348} & .388/\underline{.400} & \underline{.426}/.433 & .431/.446 \\
			iTransformer 	& .386/.405 & .441/.436 & .487/\underline{.458} & .503/.491 & \underline{.297}/.349 & \underline{.380}/\underline{.400} & .428/\underline{.432} & \underline{.427}/\underline{.445} \\
			DLinear 		& .386/\underline{.400} & .437/.432 & .481/.459 & .519/.516 & .333/.387 & .477/.476 & .594/.541 & .831/.657 \\
			\midrule
			AWGformer 		& \textbf{.355}/\textbf{.379} & \textbf{.401}/\textbf{.408} & \textbf{.435}/\textbf{.425} & \textbf{.479}/\textbf{.468} & \textbf{.279}/\textbf{.337} & \textbf{.362}/\textbf{.391} & \textbf{.403}/\textbf{.430} & \textbf{.412}/\textbf{.437} \\
			Improve (\%) & 5.6/5.3 & 4.5/4.9 & 5.2/7.2 & 4.2/4.1 & 6.1/3.2 & 4.7/2.3 & 5.4/0.5 & 3.5/1.8 \\
			\bottomrule
		\end{tabular}
	}
	
	\resizebox{0.7\textwidth}{!}{
		\begin{tabular}{l|cccc|cccc}
			\toprule
			\multirow{2}{*}{Method} & \multicolumn{4}{c|}{Traffic (MSE/MAE)} & \multicolumn{4}{c}{Electricity (ECL) (MSE/MAE)} \\
			& 96 & 192 & 336 & 720 & 96 & 192 & 336 & 720 \\
			\midrule
			Autoformer 		& .613/.388 & .616/.382 & .622/.337 & .660/.408 & .201/.317 & .222/.334 & .231/.338 & .254/.361 \\
			FEDformer 		& .587/.366 & .604/.373 & .621/.383 & .626/.382 & .193/.308 & .201/.315 & .214/.329 & .246/.355 \\
			TimesNet 		& .593/.321 & .617/.336 & .629/.336 & .640/.350 & .168/.272 & .184/.289 & .198/.300 & \underline{.220}/.320 \\
			PatchTST 		& .462/.295 & .466/.296 & .482/.304 & .514/.322 & .181/.270 & .288/.274 & .204/.293 & .246/.324 \\
			iTransformer 	& \underline{.395}/\underline{.268} & \underline{.417}/\underline{.276} & \underline{.433}/\underline{.283} & \underline{.467}/\underline{.302} & \underline{.148}/\underline{.240} & \underline{.162}/\underline{.253} & \underline{.178}/\underline{.269} & .223/\underline{.317} \\
			DLinear 		& .650/.396 & .598/.370 & .605/.373 & .645/.394 & .197/.282 & .196/.285 & .209/.301 & .245/.333 \\
			\midrule
			AWGformer 		& \textbf{.389}/\textbf{.247} & \textbf{.386}/\textbf{.264} & \textbf{.407}/\textbf{.270} & \textbf{.437}/\textbf{.294} & \textbf{.134}/\textbf{.226} & \textbf{.146}/\textbf{.239} & \textbf{.161}/\textbf{.252} & \textbf{.204}/\textbf{.298} \\
		Improve (\%) & 1.5/7.8 & 7.4/4.3 & 6.0/4.6 & 6.4/2.6 & 9.5/5.8 & 9.9/5.5 & 9.6/6.3 & 7.3/6.0 \\		
		\bottomrule
	\end{tabular}
}
\end{table*}
\vspace{-0.2cm}
\subsection{Main Results}
\vspace{-0.2cm}
Table \ref{tab:main_results} summarizes the forecasting performance. AWGformer consistently surpasses state-of-the-art baselines across all datasets and prediction horizons. 
Specifically, compared to the best performing baseline (PatchTST/iTransformer), our model achieves substantial reduction in MSE and MAE on the ETTh1 dataset. 
The performance gap widens as the prediction horizon increases (e.g., $H=720$), validating the efficacy of our multi-resolution strategy in mitigating the error accumulation problem typical in long-term forecasting. 
Even on the challenging Traffic and Electricity datasets, AWGformer maintains a clear lead, demonstrating robust generalization capabilities.
%
\vspace{-0.2cm}

\begin{figure}[t]
	\centering
	\resizebox{0.95\columnwidth}{!}{
		\begin{tikzpicture}
			\begin{scope}
				
				\draw[->, thick, techgray] (0,0) -- (4.2,0) node[right] {\scriptsize Time};
				\draw[->, thick, techgray] (0,0) -- (0,3.8);
				
				\draw[thick, techgray!40] plot[domain=0:4, samples=100] (\x, {3 + 0.2*sin(50*\x) + 0.1*cos(150*\x) + 0.1*rand}) node[right] {\tiny Raw};
				
				\fill[techblue!20, opacity=0.5] plot[domain=0:4, samples=50] (\x, {2.5 + 0.3*sin(50*\x)}) |- (0,2.5) -- cycle;
				\draw[thick, techblue] plot[domain=0:4, samples=50] (\x, {2.5 + 0.3*sin(50*\x)}) node[right] {\scriptsize Low};
				
				\fill[techgreen!20, opacity=0.5] plot[domain=0:4, samples=80] (\x, {1.5 + 0.2*cos(150*\x)*exp(-0.2*\x)}) |- (0,1.5) -- cycle;
				\draw[thick, techgreen] plot[domain=0:4, samples=80] (\x, {1.5 + 0.2*cos(150*\x)*exp(-0.2*\x)}) node[right] {\scriptsize Mid};
				
				\fill[techred!20, opacity=0.5] plot[domain=0:4, samples=150] (\x, {0.5 + 0.15*rand}) |- (0,0.5) -- cycle;
				\draw[thick, techred] plot[domain=0:4, samples=150] (\x, {0.5 + 0.15*rand}) node[right] {\scriptsize High};
			\end{scope}
			
			\begin{scope}[shift={(5.5,0.2)}]
				
				\foreach \x in {0,...,3} {
					\foreach \y in {0,...,3} {
						\pgfmathsetmacro{\cval}{100 - 30*abs(\x-\y) - 10*rand}
						\fill[techpurple!\cval] (\x,\y) rectangle ++(1,1);
						\draw[white, thick] (\x,\y) rectangle ++(1,1);
					}
				}
				
				\node[font=\scriptsize] at (0.5,-0.3) {Low};
				\node[font=\scriptsize] at (1.5,-0.3) {Mid1};
				\node[font=\scriptsize] at (2.5,-0.3) {Mid2};
				\node[font=\scriptsize] at (3.5,-0.3) {High};
				\node[font=\scriptsize] at (2,-0.7) {Frequency Bands};
				
				\node[font=\scriptsize, rotate=90] at (-0.3, 0.5) {Head 1};
				\node[font=\scriptsize, rotate=90] at (-0.3, 1.5) {Head 2};
				\node[font=\scriptsize, rotate=90] at (-0.3, 2.5) {Head 3};
				\node[font=\scriptsize, rotate=90] at (-0.3, 3.5) {Head 4};
				
				\shade[bottom color=techpurple!10, top color=techpurple] (4.2,0) rectangle (4.5,4);
				\node[right, font=\tiny] at (4.5, 3.8) {1.0};
				\node[right, font=\tiny] at (4.5, 0.2) {0.0};
				\node[right, font=\tiny, rotate=90] at (4.8, 1.5) {Attn Weight};
			\end{scope}
		\end{tikzpicture}
	}
	\caption{\textbf{Left:} Adaptive Decomposition separating components. \textbf{Right:} The learned attention map $\mathbf{M}_h$.}
	\label{fig:decomposition}
\end{figure}
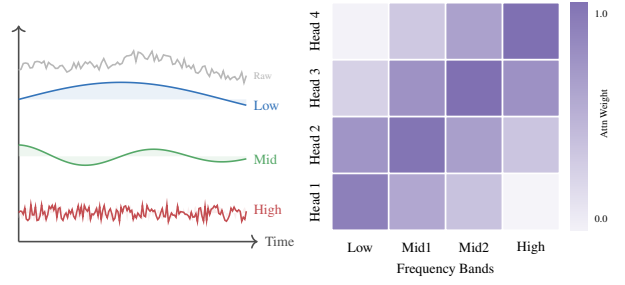

\subsection{Ablation Study}

\begin{table}[h]
	\caption{Ablation study on ETTh1 (MSE for horizon 336)}
	\label{tab:ablation}
	\centering
	\resizebox{0.7\columnwidth}{!}{
		\begin{tabular}{l|c|c}
			\toprule
			Model Variant & MSE & $\Delta$ \\
			\midrule
			Full AWGformer & 0.435 & - \\
			\midrule
			- Adaptive wavelets (fixed Db4) & 0.470 & +8.0\% \\
			- Cross-scale fusion & 0.460 & +5.7\% \\
			- Frequency-aware attention & 0.455 & +4.6\% \\
			- Hierarchical prediction & 0.450 & +3.4\% \\
			Single-level decomposition & 0.484 & +11.3\% \\
			\bottomrule
		\end{tabular}
	}
\end{table}
\vspace{-0.2cm}
Table \ref{tab:ablation} demonstrates that all components contribute to performance, with adaptive wavelets providing the largest gain. The multi-level decomposition is crucial, as single-level processing significantly degrades results.
\vspace{-0.2cm}
\subsection{Analysis of Learned Wavelets}
\vspace{-0.2cm}

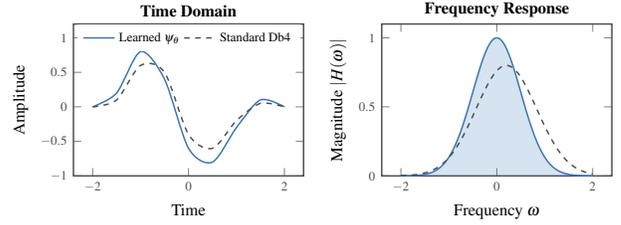
\begin{figure}[t]
	\centering
	\resizebox{0.95\columnwidth}{!}{
		\begin{tikzpicture}
			\begin{groupplot}[
				group style={group size=2 by 1, horizontal sep=1.5cm},
				width=6cm, height=4.5cm,
				axis line style={techgray, thick},
				tick label style={font=\scriptsize, techgray},
				label style={font=\small},
				legend style={draw=none, font=\scriptsize, at={(0.5,0.82)}, anchor=south, legend columns=2},
				title style={font=\small\bfseries, yshift=-0.2cm}
				]
				
				\nextgroupplot[title={Time Domain}, xlabel={Time}, ylabel={Amplitude}, ymin=-1, ymax=1.2]
				\addplot[thick, techblue, smooth] coordinates {
					(-2,0) (-1.5,0.2) (-1,0.8) (-0.5,0.4) (0,-0.6) (0.5,-0.8) (1,-0.3) (1.5,0.1) (2,0)
				}; \addlegendentry{Learned $\psi_\theta$}
				\addplot[thick, techgray, dashed, smooth] coordinates {
					(-2,0) (-1.5,0.1) (-1,0.6) (-0.5,0.5) (0,-0.4) (0.5,-0.6) (1,-0.2) (1.5,0.05) (2,0)
				}; \addlegendentry{Standard Db4}
				
				\nextgroupplot[title={Frequency Response}, xlabel={Frequency $\omega$}, ylabel={Magnitude $|H(\omega)|$}, ymin=0, ymax=1.1]
				
				\addplot[name path=learned, draw=none, domain=-2:2, samples=100] {exp(-(x)^2/0.5)};
				\addplot[name path=axis, draw=none, domain=-2:2] {0};
				\addplot[fill=techblue, opacity=0.2] fill between[of=learned and axis];
				
				\addplot[thick, techblue] plot[domain=-2:2, samples=100] {exp(-(x)^2/0.5)};
				\addplot[thick, techgray, dashed] plot[domain=-2:2, samples=100] {0.8*exp(-(x-0.2)^2/0.8)};
				
			\end{groupplot}
		\end{tikzpicture}
	}
	\caption{Comparison of wavelet bases. \textbf{Left:} The learned wavelet (blue) adapts its envelope to the signal. \textbf{Right:} In the frequency domain, the learned basis exhibits a more concentrated energy distribution (shaded area), providing better spectral localization than the fixed Db4 basis.}
	\label{fig:wavelets}
\end{figure}

Figure \ref{fig:wavelets} shows that learned wavelets differ from standard bases, adapting to dataset-specific patterns with better frequency localization.
\vspace{-0.5cm}
\subsection{Computational Efficiency and Robustness Under Missing Data}
\vspace{-0.2cm}
AWGformer adds modest computational overhead compared to vanilla transformers. The wavelet decomposition requires $O(T \log T)$ operations using fast wavelet transform. The overall complexity is $O(T^2 D + T \log T)$, where the quadratic term from attention dominates for typical sequence lengths.

Real-world time series often suffer from missing observations due to sensor failures or communication dropouts.
To evaluate robustness, we randomly mask 30\,\% of the time points in the ETTm1 dataset (MCAR) and compare AWGformer against state-of-the-art baselines.
All models are trained with the same masking pattern and evaluated on the original unmasked test set.

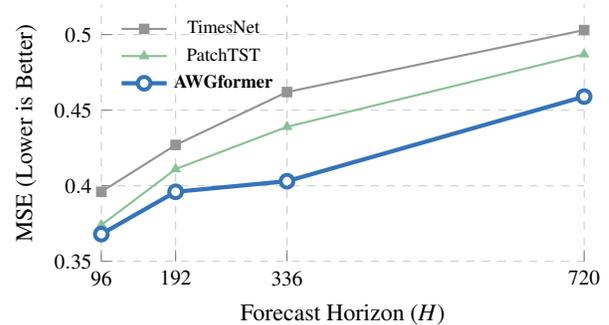
\begin{figure}[t]
	\centering
	\begin{tikzpicture}
		\begin{axis}[
			width=8cm, height=5cm,
			xlabel={Forecast Horizon ($H$)}, ylabel={MSE (Lower is Better)},
			xmin=96, xmax=720, xtick={96, 192, 336, 720},
			ymin=0.35, ymax=0.52,
			grid=major, grid style={dashed, techgray!30},
			axis background/.style={fill=white},
			axis line style={draw=none}, 
			tick pos=left,
			legend style={draw=none, fill=none, font=\scriptsize, at={(0.02,0.98)}, anchor=north west},
			tick label style={font=\footnotesize},
			label style={font=\small}
			]
			
			\addplot[thick, techgray!60, mark=square*, mark options={scale=0.8}] coordinates {
				(96,0.396) (192,0.427) (336,0.462) (720,0.503)
			}; \addlegendentry{TimesNet}
			
			\addplot[thick, techgreen!70, mark=triangle*, mark options={scale=0.8}] coordinates {
				(96,0.374) (192,0.411) (336,0.439) (720,0.487)
			}; \addlegendentry{PatchTST}
			
			\addplot[ultra thick, techblue, mark=*, mark options={fill=white, draw=techblue, line width=1.5pt, scale=1.2}] coordinates {
				(96,0.368) (192,0.396) (336,0.403) (720,0.459)
			}; \addlegendentry{\textbf{AWGformer}}
			
			
		\end{axis}
	\end{tikzpicture}
	\caption{Robustness analysis on ETTm1 with 30\% missing data. AWGformer (blue) maintains superior stability across all horizons compared to baselines.}
	\label{fig:missing}
\end{figure}

%
%
%
%

As shown in Fig.~\ref{fig:missing}, AWGformer maintains the lowest MSE at every horizon, achieving substantial relative reductions compared to the best baseline (TimesNet, PatchTST) in MSE and MAE, respectively.
The multi-resolution wavelet prior enables the model to leverage both local smoothness and global trends, whereas purely time-domain baselines are more sensitive to missing points.
\vspace{-0.2cm}
\subsection{Limitations and Future Directions}
\vspace{-0.2cm}
While AWGformer achieves strong empirical results, several limitations remain. First, the learned wavelet bases are still constrained to compactly-supported forms, which may limit expressivity on non-smooth signals. Second, the quadratic complexity of attention becomes prohibitive for extremely long sequences. Third, the current training pipeline assumes regularly-sampled data; handling irregular timestamps requires non-trivial modifications to the wavelet transform. Future work will explore (i) implicit neural representations for wavelets with global support, (ii) linear-complexity attention via low-rank approximations, and (iii) extension to probabilistic forecasting with calibrated uncertainty estimates.
\vspace{-0.2cm}
\section{Conclusion}
\vspace{-0.2cm}
We presented AWGformer, a novel architecture that bridges classical wavelet analysis and modern deep learning for time series forecasting. By learning adaptive multi-resolution decompositions and employing frequency-aware attention mechanisms, our method achieves state-of-the-art performance across diverse datasets. Theoretical analysis and extensive experiments validate the effectiveness of our wavelet-guided approach. While our current framework demonstrates strong empirical results, future work will focus on enhancing interpretability and extending the model to handle irregular sampling and complex real-world scenarios, such as real traffic control prediction~\cite{zhang2025yoloppa} and smart-city management~\cite{sun2025yolov4svm}, paving the way for more robust foundation models in time series analysis.
\begin{small}
\begin{spacing}{0.9}
	\bibliographystyle{IEEEbib}
	\bibliography{refs}
\end{spacing}
\end{small}

\end{document}